\documentclass[letterpaper, 10 pt, conference]{ieeeconf}  % Comment this line out if you need a4paper

\IEEEoverridecommandlockouts                              % This command is only needed if 
\usepackage{epsfig} % for postscript graphics files

\usepackage[english]{babel}
\usepackage{graphicx}
\usepackage{amsmath, amsfonts, amssymb}
\usepackage{siunitx}
\usepackage{caption}
\usepackage{subcaption}
\usepackage{color}

\title{\LARGE \bf
Curriculum-based Sample Efficient Reinforcement Learning for Robust Stabilization of a Quadrotor
}

\author{Fausto Mauricio Lagos Suarez$^{1}$, Akshit Saradagi, Vidya Sumathy, \\ Shruti Kotpalliwar and George Nikolakopoulos% <-this % stops a space
\thanks{*This work has been funded by the European Union's Horizon Europe Research and Innovation Program, under the Grant Agreement No. 101119774 SPEAR.}% <-this % stops a space
\thanks{*This research was conducted using the resources of High Performance Computing Center North (HPC2N). Additionally, the RL-training were enabled by resources provided by the National Academic Infrastructure for Supercomputing in Sweden (NAISS), partially funded by the Swedish Research Council through grant agreement no. 2022-06725.}
\thanks{$^{1}$Fausto Lagos is the corresponding author of the article
        {\tt\small faulag@ltu.se}. The authors are with the Robotics and AI group, in the Department of Computer Science, Electrical and Space Engineering at Luleå University of Technology, Sweden.}%
}

\begin{document}

\maketitle
\thispagestyle{empty}
\pagestyle{empty}

\begin{abstract}
This article introduces a novel sample-efficient curriculum learning (CL) approach for training an end-to-end reinforcement learning (RL) policy for robust stabilization of a Quadrotor. The learning objective is to simultaneously stabilize position and yaw-orientation from random initial conditions through direct control over motor RPMs (end-to-end), while adhering to pre-specified transient and steady-state specifications. This objective, relevant in aerial inspection applications, is challenging for conventional one-stage end-to-end RL, which requires substantial computational resources and lengthy training times. To address this challenge, this article draws inspiration from human-inspired curriculum learning and decomposes the learning objective into a three-stage curriculum that incrementally increases task complexity, while transferring knowledge from one stage to the next. In the proposed curriculum, the policy sequentially learns hovering, the coupling between translational and rotational degrees of freedom, and robustness to random non-zero initial velocities, utilizing a custom reward function and episode truncation conditions. The results demonstrate that the proposed CL approach achieves superior performance compared to a policy trained conventionally in one stage, with the same reward function and hyperparameters, while significantly reducing computational resource needs (samples) and convergence time. The CL-trained policy's performance and robustness are thoroughly validated in a simulation engine (Gym-PyBullet-Drones), under random initial conditions, and in an inspection pose-tracking scenario. A video presenting our results is available at https://youtu.be/9wv6T4eezAU.
\end{abstract}

% \keywords{Curriculum Learning, Reinforcement Learning, Quadrotor, Stabilization, Trajectory Tracking} 

%===============================================================================

\section{Introduction}
In the recent years, Reinforcement Learning (RL) has emerged as an appealing control design tool for engineers, with and without formal training in machine learning and automatic control. This is especially true in the design of autonomy for Quadrotors, where RL is being used to learn policies for path planning, navigation, and control \cite{alvarez_forest_2023}. 
%\cite{alvarez_forest_2023}. 
RL has proven particularly effective in tackling complex aerial control tasks that are challenging for classical system-theoretic approaches \cite{antonyshyn_deep_2024}. In RL, an agent learns a specific behavior in a model-free fashion, by interacting with its environment, by taking actions, and receiving numeric rewards that indicate the effectiveness of the actions in achieving the desired behavior \cite{sutton_reinforcement_2018}. This makes RL well-suited for complex learning tasks involving UAVs, where precise modeling of the interaction between the UAV's nonlinear dynamics and the surrounding aerial flows is highly challenging, especially for soft and unconventional drone designs \cite{zhang_extreme}. 

Although recent RL demonstrations portray RL to be almost akin to plug-and-play, researchers in academia and industry recognize the significant human time and effort involved in solving complex control problems using RL, even when powerful computing clusters are employed. It is not uncommon for even experienced Ph.D students in academia and trained personnel in industry to spend several months in iteratively perfecting RL training, with reward design and hyperparameter optimization accounting for a major part of the time and effort. Although using modern GPUs and extreme parallelization in training, millions of time steps (samples) of training can be achieved in minutes \cite{eschmann_learning_2024} \cite{kulkarni_aerial_2023}, RL training for UAVs is still computationally intensive and time-consuming, which underscores the need to investigate the fundamental challenge of sample-efficiency in order to develop sample-efficient RL training methods. 
%
%The control objective under consideration and the training methodology aim to advance the use of RL in developing robust, high-accuracy low-level controllers for Quadrotors, contrasting with
%
% To add, 

Recent literature on RL-based control for Quadrotors has explored diverse aspects: design of Gym environments for parallelized training \cite{kulkarni_aerial_gym} \cite{dimmig_aerial_survey}, training setups for zero-shot sim-2-real transfer \cite{gronauer_zero_shot}, exploration of the right action and observation spaces \cite{kaufman_comparison} \cite{dionigi_inputs}, design of reward functions to achieve specific aerial objectives \cite{ganai_reward}, generalization of RL training to multiple Quadrotors through domain randomization \cite{eschmann_raptor}, improvements to training algorithms such as the Proximal Policy Optimization (PPO) algorithm for improved learning stability \cite{xue_improved_2022}, exploration of different training algorithms such as DDPG, TRPO, PPO for quadrotor control \cite{bernini_few_2021} etc.  
%\cite{lillicrap_continuous_2019}, Trust Region Policy Optimization (TRPO) \cite{schulman_trust_2017}, and PPO \cite{schulman_proximal_2017} etc. 
Additionally, \cite{hwangbo_control_2017} introduced an RL-based controller capable of stabilizing a Quadrotor from extreme initial conditions, such as manual tossing, which is relevant in the context of this article.  

\begin{figure*}[t]
    \centering
     \includegraphics[width=\linewidth]{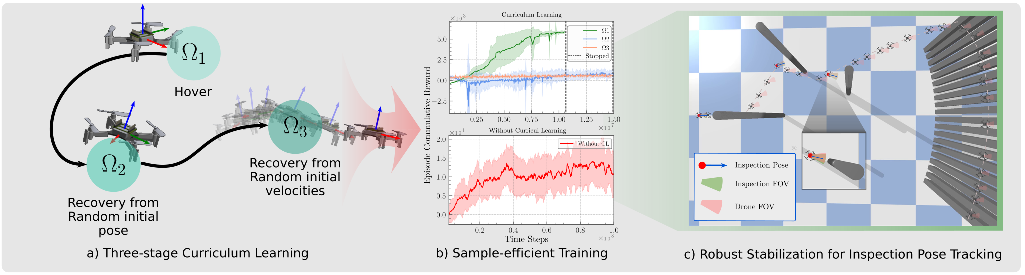} 
    \caption{A curriculum learning-inspired computationally-efficient three-stage RL training approach, to learn a policy for quadrotor control, while meeting pre-specified transient and steady-state requirements from aerial inspection missions.}
    \label{fig:concept}
\end{figure*}

It is worth noting that, in the literature alluded to so far, a conventional one-stage training approach has been utilized, with no mention of the sample efficiency of the training process and the significant computation and wall-clock time spent in tuning the RL training. In this context, this article draws inspiration from bio-inspired curriculum learning and proposes a light, structured and modular three-stage RL-training workflow to achieve robust stabilization for a Quadrotor (position and yaw) starting from random initialization in position, orientation, and non-zero linear and angular velocities. This learning objective, highly relevant in aerial inspection and monitoring applications, is challenging for single-stage training commonly used in RL literature, which requires several hundred million time steps in one-stage training \cite{xu_omnidrones_2023}.  Curriculum learning, which has gained popularity in legged robotics \cite{feng_genloco} and autonomous robotics \cite{beigomi_autonomous}, has not yet been exploited in RL-based quadrotor control literature. This is mainly due to the emergence of GPU-aided Gyms, which have significantly reduced the time between two tuning runs. 
%It is only in [learning to fly in seconds] that a primitive version has been explored, as an iterative update of the weights of the reward function. 
%===============================================================================

\textit{Contributions:} Given the premise, the contributions of this article (depicted pictorially in Fig. \ref{fig:concept}) are presented next. This work presents a novel three-stage Curriculum Learning (CL) methodology for training an RL agent to achieve robust stabilization of a Quadrotor (position and yaw). The focus on sample-efficient training and the goal of achieving pre-specified settling time and steady-state requirements sets this article apart from existing literature. 1) The proposed curriculum (Section \ref{sec:methodology}), wherein the domain of quadrotor initializations is progressively expanded in three stages, is designed while taking into account the underactuated nature of the Quadrotor and the coupling between the translational and rotational degrees-of-freedom. This is unlike in \cite{eschmann_learning_2024}, wherein the curriculum is set up to sequentially change the weights in the reward functions.
2) This work proposes a reward function and episode truncation conditions to achieve pre-specified settling time and steady-state requirements, motivated by the needs of the aerial inspection and monitoring applications.
3) The results (Section \ref{sec:training}) demonstrate that the proposed CL-training significantly enhances the sample efficiency of the training process compared to single-stage training, which fails to achieve the considered learning objective, even with extended training periods, with the same reward function and hyperparameters of the training algorithm. The robustness of the CL-trained policy is validated through extensive testing from a diverse set of initial conditions and in an inspection pose tracking scenario.
\section{Problem Formulation}
\begin{figure}[h!]
    \centering
    \includegraphics[scale=0.8]{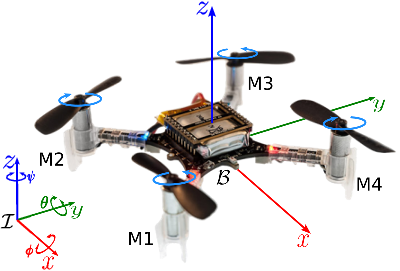}
    \caption{The Crazyflie Quadrotor.}
    \label{fig:quadrotor_bodyframe}
\end{figure}

A Quadrotor is an underactuated aerial system with six degrees-of-freedom (DOF): three translational ($(x, y, z)\in\mathbb{R}^3$) and three rotational ($\phi, \theta, \psi \in \mathbb{S}^1 \times \mathbb{S}^1 \times \mathbb{S}^1$). The Quadrotor is influenced by four control inputs supplied to the four motors M1-M4. 
The motors produce upward thrust and the roll, pitch, and yaw rotations necessary for aerial mobility. In this article, a Crazyflie 2.1 (shown in Figure \ref{fig:quadrotor_bodyframe}) in $\times$ configuration is considered. Mathematical modeling of the Quadrotor dynamics can be found in \cite{luis_design_2016}. %\cite{beard_quadrotor_nodate}
\textbf{Problem statement.} \label{sec:problem_statement} The training of an end-to-end RL policy to achieve robust stabilization of a Quadrotor, with acceptable performance levels, requires millions of interactions \cite{ferede_end--end_2024} with the training environment, despite the availability of Gym environments aided by high-performance Graphics Processing Units (GPUs). 
%This high demand for interactions makes the training process computationally expensive and time-consuming. 
To address this challenge, this work considers the problem of developing a methodology for RL training that can be a sample-efficient alternative to the ubiquitous one-stage training. The learning objective is to achieve robust stabilization of a Quadrotor through direct control of the Quadrotor's RPMs (end-to-end), from random initial states (including non-zero velocities). 
Driven by the rigorous demands on Quadrotor stability in aerial inspection missions, the objective also includes satisfying desired control-theoretic performance metrics such as settling time and steady-state accuracy in both position and yaw-orientation. For validation, the performance specifications are set to: i) settling time of less than 5 seconds, ii) positional steady-state error within 2.5 cm of the target position, and iii) yaw-tracking error within $2^\circ$ of the commanded target yaw orientation.
%-------------------------------------------
\section{Curriculum Learning Methodology} \label{sec:methodology}
Curriculum learning involves three key components: sub-task generation, sequencing, and transfer learning. This approach decomposes a complex target task into a series of sub-tasks, progressively transferring knowledge through multiple learning stages until the target is achieved \cite{narvekar_curriculum_2020}. This section introduces a sequenced three-stage curriculum, designed to enhance sample efficiency of the RL training process for Quadrotor control. 
In synthesizing a sequence of three sub-tasks, the Quadrotor's under-actuated nature and the coupling between different degrees-of-freedom of a Quadrotor are taken into account. The training initially focuses on the task of achieving stable hovering from a fixed position. The task difficulty gradually increases in two additional stages, where random initialization in positions/orientations and velocities (both linear and angular), respectively, are introduced along with the requirement for achieving $0^\circ$ as target yaw. In general, sub-tasks may differ from the final task in terms of state/action space, reward function, or transition dynamics \cite{karlsson_task_nodate} \cite{kwon_reinforcement_2024}. In this work, we maintain a consistent reward function structure across all sub-tasks, but sequentially expand the domain of initialization of the Quadrotor, thereby altering the environment within the Markov Decision Process (MDP). 
The \textbf{first sub-task ($\Omega_1$)} focuses on achieving a target hover position at $(x(t),y(t),z(t))=(0, 0, 1)$, starting from the fixed position $(x(0),y(0),z(0))=(0, 0, 0)$. In this stage, the policy is trained to control the drone to take off and maintain the target position, which essentially involves learning to use the same RPM values for all four motors. Such a policy is better suited to learning the sub-task $\Omega_2$ presented next, compared to a policy that is randomly initialized (as in one-stage training). 
The \textbf{second sub-task ($\Omega_2$)} increases task complexity by requiring the drone to reach $(x(t),y(t),z(t))=(0, 0, 1)$ and $\psi(t) = 0$, from random initial positions within a cylinder of 2 meters radius and 2 meters height, and random initial attitudes with roll and pitch angles in a safe range of $[-15^\circ, 15^\circ]$ and yaw angle in a range of $[-180^\circ, 180^\circ)$. This task emphasizes learning the coupling between the $x$ and $y$ degrees-of-freedom and the roll $(\phi)$ and pitch $(\theta)$ rotations. A policy that achieves task $\Omega_2$ is well placed to learn the final task $\Omega_3$ presented next, compared to a policy that is randomly initialized (as in one-stage training).
The \textbf{third sub-task ($\Omega_3$}), which corresponds to the learning objective of robust stabilization, further increases difficulty by introducing random initial linear velocities within the range of $[-1, 1]$ meters per second and angular velocities within the range of $[-1, 1]$ radians per second, in addition to random initial positions and orientations. Here, the policy learns to command the four motors to achieve the desired position and yaw, while correcting for deviations induced by random initial velocities.
\begin{figure}[h!]
    \centering
    \includegraphics[width=\linewidth]{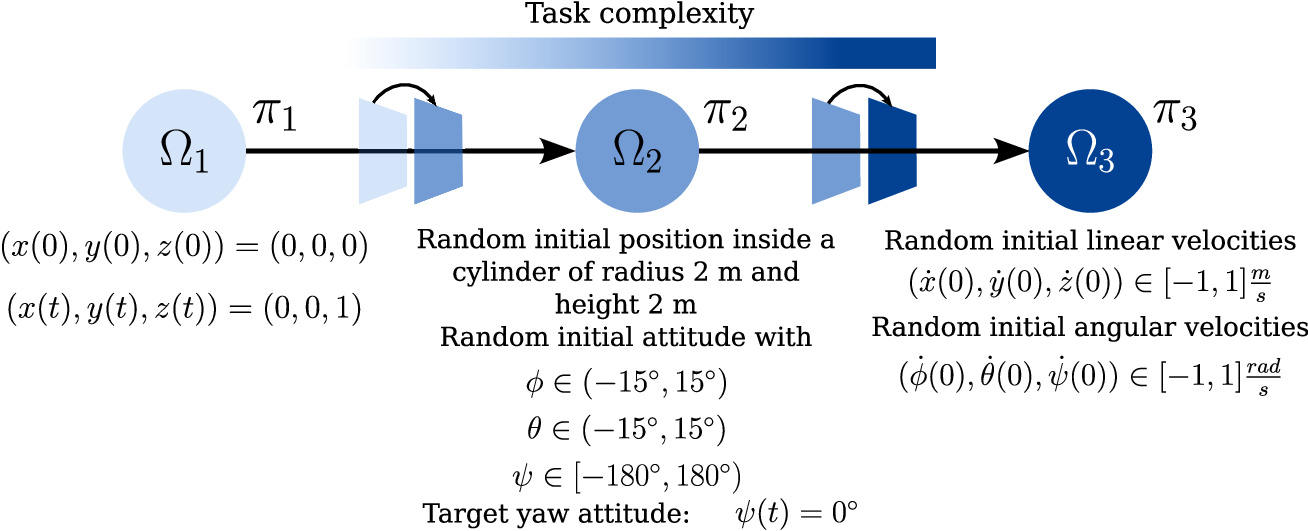}
    \caption{The sequence of tasks in the proposed three-stage curriculum learning, showing the evolution of the task complexity and variations in the domain of initialization.}
    \label{fig:cl_graph}
\end{figure}
A pictorial illustration of the three-stage curriculum is presented in Fig. \ref{fig:cl_graph}. To evaluate the gains in sample efficiency and convergence time of the proposed curriculum learning against the baseline of one-stage training, the number of time steps and wall-clock time required to achieve the target task are used as metrics in Section \ref{sec:training}. The evaluation focuses on showcasing the ability of the curriculum-trained policy to achieve the learning objective while yielding significant reduction in training time and enhanced sample efficiency.
\section{Reinforcement learning setup} \label{sec:RL}
In the reinforcement learning framework, the control of a Quadrotor is modeled as a Markov Decision Process and represented as a tuple of five elements $(\mathbb{E}, \pi, \mathbb{A}, \mathbb{S}, \mathbf{R})$, where $\mathbb{E}$ is the environment (the simulation engine \textit{Gym-PyBullet-Drones} in this case), $\mathbb{A}$ is the set of actions, $\mathbb{S}$ and $\mathbf{R}$ are the current state (observation) and rewards returned by the environment respectively, and $\pi$ is the policy (a neural network) in charge of taking control decisions (mapping observations to actions). 
\subsection{Observation and action spaces}
In this work, we consider a 12-dimensional observation space consisting of the position $[\begin{array}{ccc}x & y & z\end{array}] \in \mathbb{R}^3$, the Euler angles $[\begin{array}{ccc}\phi & \theta & \psi\end{array}] \in \mathbb{S}^1 \times \mathbb{S}^1 \times \mathbb{S}^1$, the linear velocities $[\begin{array}{ccc}\dot{x} & \dot{y} & \dot{z}\end{array}] \in \mathbb{R}^3$ and the angular velocities $[\begin{array}{ccc}\dot{\phi} & \dot{\theta} & \dot{\psi}\end{array}] \in \mathbb{R}^3$. The rotor speeds $R_1, \ldots, R_4$ (in RPM) of the propellers constitute the action space. The end-to-end reinforcement learning policy $\pi$ directly maps the observation space (states) to the action space (RPMs of the motors). The Crazyflie Quadrotor considered in this work allows for direct control with the RPMs of the motors.
\subsection{Reward shaping} \label{sec:reward_shaping}
The goal of an RL algorithm is to train an agent to obtain the most discounted cumulative reward over time. Although several works in literature (such as \cite{jiang_quadrotor_2021}) suggest having a simple reward function based on just the error in position and velocities, simple rewards lead to sample inefficiency, as the policy explores without adequate feedback, and are not suitable for problems that go beyond hovering. To develop an RL policy that yields high accuracy, stability, and robustness, while having good sample efficiency in training, a reward function that encodes such expectations must be crafted. In this article, we design an additive reward function that penalizes excessive exploration, instability, and imprecision in reaching the target position and attitude, that are crucial for both robust stabilization and improved sample efficiency of the training process. In this article, we propose the following reward function:
\begin{equation}
    R(t) = 25 - 20T_e - 100E + 20S - 18w_e,
    \label{eq:reward_function}
\end{equation}
where the individual components are defined below. Although some of the terms are in fact penalties and not rewards, we use the term 'reward' without loss of generality.   

\textbf{Target reward $(T_e)$:} The target reward is a penalty proportional to the distance between the Quadrotor and the target position ($T$), and the difference between the target yaw and the current yaw. The penalty decreases as the Quadrotor nears the target configuration. 
The penalty is computed as:
\begin{equation} \label{eq:target_reward}
    T_e = ||[\begin{array}{ccc} x_T & y_T & z_T\end{array}] - [\begin{array}{ccc} x & y & z\end{array}]|| + |\psi_T - \psi|,
\end{equation}
where the targets are indicated through a subscript. 

\textbf{Exploration reward ($E$):} The exploration space for the policy is designed to be a cylinder, centered at the target position ($T$). In each episode, the cylinder's radius is set to the distance between the center and the starting position of the quadrotor plus a tolerance $\delta_R$, and the height is the target height $z$ plus a tolerance $\delta_H$. This bounded exploration space, along with the truncation conditions for each training episode, helps in lowering the overall training time, without compromising on the agent's ability to explore adequately. The exploration reward ($E$) is defined as:
\begin{equation} \label{eq:exploration_reward}
    E = \begin{cases}
        1 & d(C, T) > d(i, T) + \delta_R \vee C_z > T_z + \delta_H \\
        -0.2 & \text{otherwise}
    \end{cases},
\end{equation}
where $i$, $C$ and $T$ refer to the initial, current and target positions, respectively.

\textbf{Stability reward ($S$):} To ensure that the drone achieves stability at the target position, we provide a positive reward when the drone is within $\Delta_p$ cm of the target position and the sum-of-squares of the roll and pitch angles are within $\Delta_a$. Otherwise, the drone receives a penalty proportional to the sum of deviations in roll and pitch from zero. The stability reward ($S$) is defined as:
\begin{equation} \label{eq:stability_reward}
    S = \begin{cases}
        2 & d(C, T) < \Delta_p \wedge \phi^2 + \theta^2 < \Delta_a \\
        -(\phi^2 + \theta^2)
    \end{cases}
\end{equation}
The parameter $\Delta_p$ represents the radius of a tolerance sphere centered at the target position and helps in imposing a pre-specified steady-state error, while $\Delta_a$ sets the tolerance on the final roll ($\phi$) and pitch ($\theta$) angles.

\textbf{Navigation reward ($w_e$):} To ensure smooth navigation from the starting position to the target position, we penalize significant changes in angular velocities. The penalty is proportional to the difference between the angular velocity in the previous time step and the current time step. 
The navigation reward ($w_e$) is calculated as 
\begin{equation}
    w_e = \sum_{i=1}^3 (w_{i_{t-1}} - w_{i_{t}})^2,
\end{equation}
where $w \in \{\dot{\phi}, \dot{\theta}, \dot{\psi\}}$. 

\textbf{Episode length and Episode truncations}: The episode length is set to the settling time specified in the performance requirements (5 seconds in the results). The stability reward in Equation \ref{eq:stability_reward} is designed to incentivize the policy to maintain Quadrotor stability beyond the episode length. In Equation \ref{eq:exploration_reward}, $\delta_R = 2.5$ cm is chosen as a tolerance margin for the radius of the cylinder. If at any moment of the episode, the distance to the target overshoots the radius of the cylinder plus $\delta_R$, the episode is truncated. Although in the reward function, specific penalties for high roll or pitch angles are not imposed, episodes are truncated when the roll or pitch angle exceeds $\pm 15^\circ$.
%------------------------------------
\section{Procedure for Sequential Training} \label{sec:training}
In this article, the standard implementation of the Proximal Policy Optimization (PPO) algorithm, considered a state-of-the-art RL algorithm \cite{xue_improved_2022}, is used to train the curriculum learning policy. 
\renewcommand{\tabcolsep}{2.5mm}
\renewcommand{\arraystretch}{1}
\renewcommand{\arrayrulewidth}{1pt}
\begin{table}[h!]
    \small{
    \centering
    \begin{tabular}{cc|cc}
        \hline
        \textbf{Policy} & MlpPolicy & \textbf{Epochs} & 10 \\
        \textbf{Learning rate} & \num{3d-4} & \textbf{Discount Factor} ($\gamma$) & 0.99\\
        \textbf{Batch Size} & 128 & \textbf{Episode Length} & 5 s\\
        \hline
    \end{tabular}
    \caption{Parameters for the PPO algorithm.}
    \label{tab:PPO_parameters}}
\end{table}

\renewcommand{\tabcolsep}{1mm}
\renewcommand{\arraystretch}{1}
\renewcommand{\arrayrulewidth}{1pt}
\begin{table}[h!]
    \centering
    \begin{tabular}{ccc}
        \hline
        \textbf{Parameter} & \textbf{Notation} & \textbf{Value} \\
        \hline
        Mass &$m$ & 0.027 [$Kg$] \\
        Arm length &$d$ & \num{39.73d-3} [$m$] \\
        Propeller radius &$p$ & \num{23.1348d-3} [$m$] \\
        Moment of Inertial about $x$ axis &$I_{xx}$ & \num{1.395d-5} [$Kg \times m^2$] \\
        Moment of Inertial about $y$ axis &$I_{yy}$ & \num{1.436d-5} [$Kg \times m^2$] \\
        Moment of Inertial about $z$ axis &$I_{zz}$ & \num{2.173d-5} [$Kg \times m^2$] \\
         \hline \\
    \end{tabular}
    \caption{Physical parameters of the Crazyflie 2.x.}
    \label{tab:crazyflie_parameters}
\end{table}
\noindent \textbf{Training process and simulation engine.} Table \ref{tab:PPO_parameters} presents the parameters used in the PPO algorithm and Table \ref{tab:crazyflie_parameters} presents the physical parameters for Crazyflie 2.x Quadrotor used for simulation. 

For the training process, we used \textit{Gym-PyBullet-Drones} \cite{noauthor_utiasdslgym-pybullet-drones_nodate}, a gymnasium \cite{noauthor_gymnasium_nodate} environment that uses PyBullet Physics \cite{erwin_coumans_and_yunfei_bai_pybullet_2024} as the physics engine and StableBaselines3 \cite{raffin_stable-baselines3_2021} as the library of reliable implementations of RL algorithms in PyTorch. \textit{Gym-PyBullet-Drones} was developed to work directly with Crazyflie 2.0 in the $\times$ and $+$ configurations, and allows for introducing obstacles into the training environment and during policy testing.

\begin{figure}[h!]
    \centering
    \includegraphics[width=\linewidth]{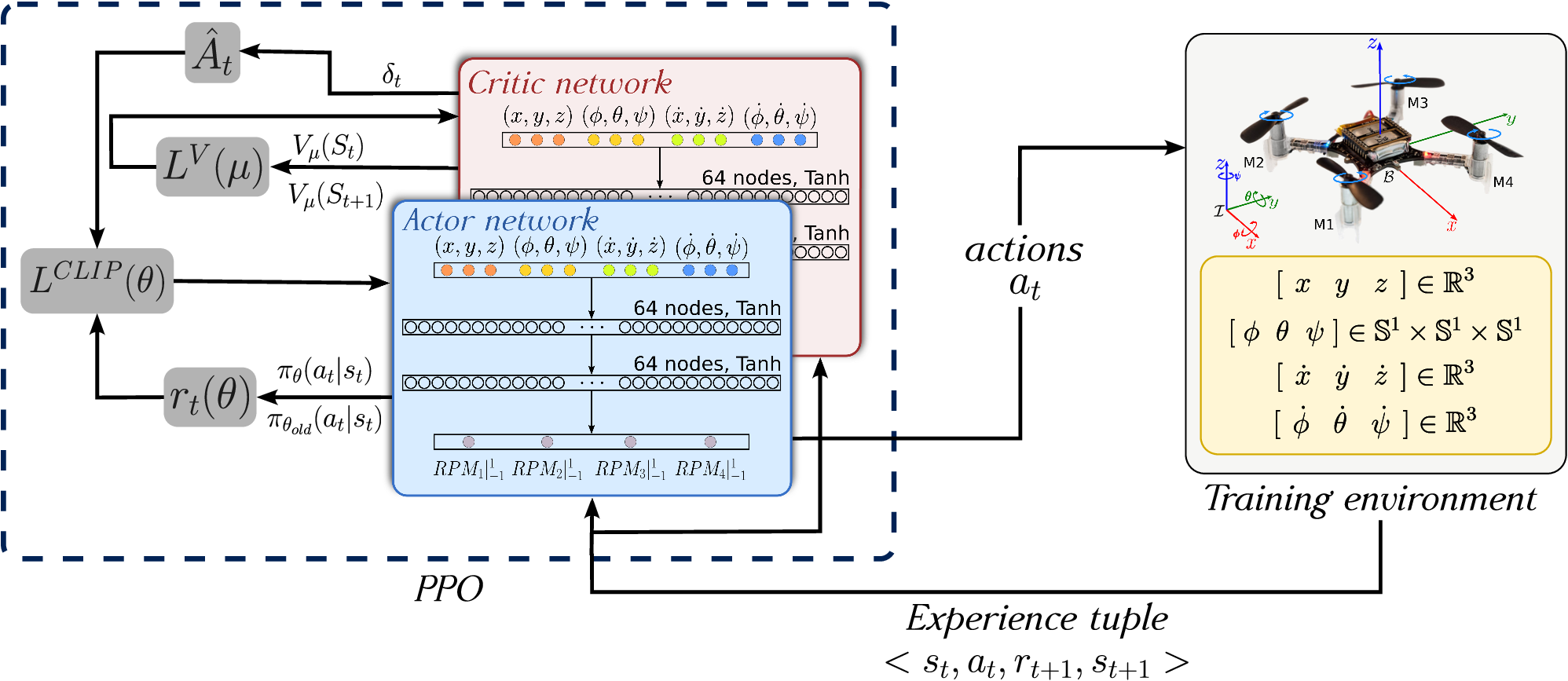}
    \caption{Reinforcement Learning setup and configuration of the Neural Networks. The RL agent (actor network) interacts with the training environment by generating actions $a_t$, based on the current state $s_t$, provided by the environment. The critic network estimates the value function, $V(s_t)$, to evaluate the state. The experience tuple $<s_t, a_t, r_{t+1}, s_{t+1}>$ is used to update both actor and critic networks through PPO's objective function, including the clipped surrogate loss, $L_{CLIP}(\theta)$.}
    \label{fig:NN_structure}
\end{figure}

The actor (policy) neural network is a Multi-layered Perceptron (MLP) network \cite{popescu_multilayer_2009} with four layers: the first is a layer of 12 inputs (the observation space), the two hidden layers have 64 fully connected nodes with $\tanh$ activation functions, and the last layer is a four-node layer (the action space). The critic (value function) network has the same architecture, except for the last layer, which is a one-node layer yielding the output of the value function. Figure \ref{fig:NN_structure} shows the RL training architecture, along with the configuration of the actor-critic neural networks used in this work. The training was executed on a single CPU on a Compute-skylake Kebnekaise HPC2N node with an Intel Xeon Gold 6132 CPU and 192 GB RAM running 4 parallel environments.
%The state of the Quadrotor includes position, orientation, linear and angular velocities.
%The actor-network outputs motor RPM values for the Quadrotor's four motors.

\begin{figure}[h!]
    \centering
    \begin{subfigure}[t]{\linewidth}
        \centering
        \includegraphics[width=.9\linewidth]{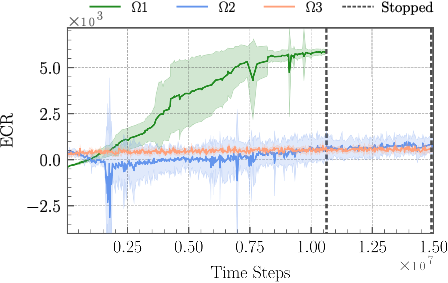}
        \caption{Three-stage CL}
    \end{subfigure}
    \quad
    \begin{subfigure}[t]{\linewidth}
        \centering
        \includegraphics[width=.9\linewidth]{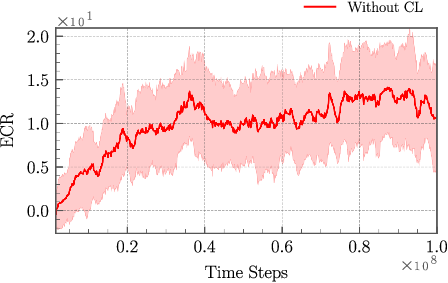}
        \caption{One stage training}
    \end{subfigure}
    \caption{Episode Cumulative Reward (ECR) comparison between curriculum learning (a) and single-stage training (b).}
    \label{fig:ECR}
\end{figure}

To demonstrate the effectiveness of the curriculum learning methodology in achieving the target task and to compare the performance with single-stage training, two agents were trained. 
%using different approaches. 
The first agent was trained directly on the target task in a single stage (without curriculum learning), and the training process was monitored using the Episode Cumulative Reward (ECR) metric. Using the training setup from Section V, after 100 million time steps, which corresponds to more than 23 hours of computation, single-stage training failed to achieve the target task, which can be inferred from the consistently low ECR achieved by single-stage training (Figure \ref{fig:ECR}). 

The second agent was trained using the three-stage curriculum learning approach proposed in this article, which achieved the target task at the end of $\Omega_3$.  The training process in each stage was monitored and stopped once the ECR stabilized for at least one million time steps. In Sub-task $\Omega_1$, the agent achieves higher cumulative rewards quickly, indicating excellent performance in learning to hover. This knowledge is then effectively transferred to subsequent tasks (Sub-task $\Omega_2$ and $\Omega_3$), which naturally exhibit lower cumulative rewards due to increased task difficulty introduced by random initial conditions and episode truncations. Table \ref{tab:training_time} presents a comparison of the time steps and wall-clock time required for single-stage training versus curriculum learning. Figure \ref{fig:ECR} provides a visual comparison of the ECR achieved by the policy trained in one stage versus the policy trained using curriculum learning.

Both the curriculum learning and single-stage training approaches were trained 
%for an equivalent duration, 
utilizing the same neural network architecture and reward function structure. The evolution of training depicted in Figure \ref{fig:ECR} and the statistics presented in Table \ref{tab:training_time} clearly show the effectiveness of curriculum learning in improving sample efficiency.
%, for the complex task of stabinvolving random initial linear and angular velocities. 
Curriculum learning allowed the agent to progressively learn increasingly difficult stabilizing maneuvers, resulting in faster convergence and higher cumulative rewards compared to single-stage training.
%-------------------------------------- 
\section{Testing of the curriculum-trained RL policy} \label{sec:results}
%

% To evaluate the performance of an agent trained with the proposed curriculum learning methodology. 

% The first was trained in achieving the target task from scratch in a single-stage and the second was trained using the proposed curriculum learning methodology. The \textbf{target task} was to achieve a fixed position at $[\begin{array}{ccc} 0 & 0 & 1\end{array}]$, starting from a random position in a cylinder of radius $r = 2$ meters and height $h = 2$ meters with random starting attitude and velocities. 
%
% Even after training for 30 million time steps, the sample efficiency for agent A was poor and the policy failed to accomplish the target task. To simplify the task for one-stage training, the requirement of random initiation of the drone's attitude was removed and the resulting agent B achieved the target task. Table \ref{tab:training_time} shows the length in time steps and hours for each training process. Each training was monitored and stopped when good stabilization at the target position was achieved (when high rewards were reached). Figure \ref{fig:ECR}, presents the evolution of the Episode Cumulative Reward (ECR) for each sub-task.
%
\begin{table}[h!]
    \small{
    \centering
    \begin{tabular}{ccc}
        \hline
        \textbf{Agent} & \textbf{Time steps} & \textbf{Wall clock} \\
        \hline
        One Stage & 100 million & 23.4 hours \\
        \hline
        % B - from scratch & 18 million & 3.6 hours \\
        Sub-task $\Omega_1$ & 11.5 million & 3 hours \\
        Sub-task $\Omega_2$ & 15 million & 4 hours \\
        Sub-task $\Omega_3$ & 15 million & 4 hours \\
        \hline
        CL $\Omega_1 \rightarrow \Omega_3$ & 41.5 million & 11 hours \\
        \hline \\
    \end{tabular}
    \caption{Sample efficiency of Curriculum Learning vs one-stage training w.r.t time steps and wall clock time.}
    \label{tab:training_time}
    }
\end{table}
% \begin{figure}[h!]
% 	\centering
%     	\includegraphics[width=0.45\textwidth]{images/training.eps}
%         \caption{}
%     \label{fig:ECR}
% \end{figure} 
%
The robustness and stability achieved by the curriculum-trained policy, is evaluated in two test scenarios: i) In the first, the performance of the policy in achieving a target position and orientation, starting from randomized position, orientation, and velocities is evaluated and ii) In the second, the accuracy in an inspection scenario, where the drone is required to reach a sequence of inspection view poses (positions with a specific orientation) is evaluated.
\begin{figure*}[h!]
    \centering
    \includegraphics[width=0.9\linewidth]{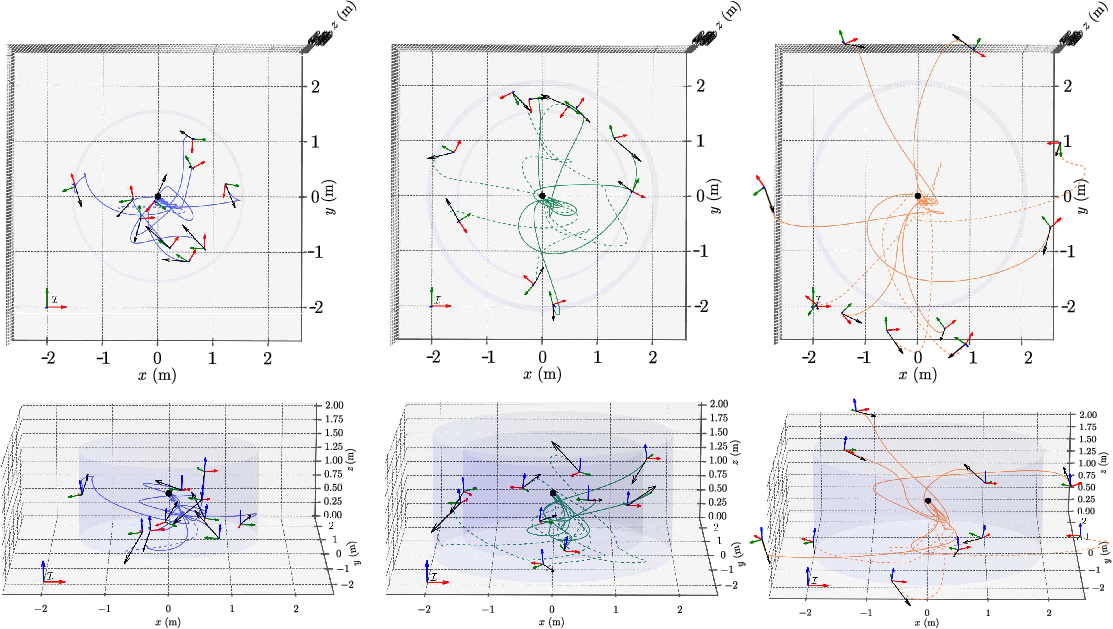}
    \caption{Evaluation of the curriculum-trained policy over 30 trials with initial positions sampled from three regions. At each initial quadrotor position, the body frame indicates attitude and a black arrow denotes the randomized initial linear velocity, with length proportional to its magnitude; random nonzero angular velocities are also applied. Solid trajectories correspond to smooth maneuvers with small transients, whereas dashed trajectories indicate large transients that occur near the ground when compensating for large unfavorable initial velocities. In all cases, the drone reaches the target position.}
    % \caption{Evaluation of the curriculum-trained policy in 30 trials, with initial positions chosen from three regions. At the starting position of the Quadrotor, we use the body coordinate frame to show variations in the initial attitude and a velocity vector (black arrow) to represent the randomized initial linear velocity (the length of the arrow is proportional to the magnitude of the velocity). The Quadrotor is also initialized with random non-zero angular velocities. Solid trajectories represent smooth maneuvers with small transients, while dashed trajectories are used to indicate large transients, which are expected when the drone gets close to the ground, to compensate for large, unfavorable initial velocities. In all tests, the drone successfully reached the target position.}  
    \label{fig:robust_stabilization}
\end{figure*}
\begin{figure*}[ht!]
    \centering
    \includegraphics[width=\linewidth]{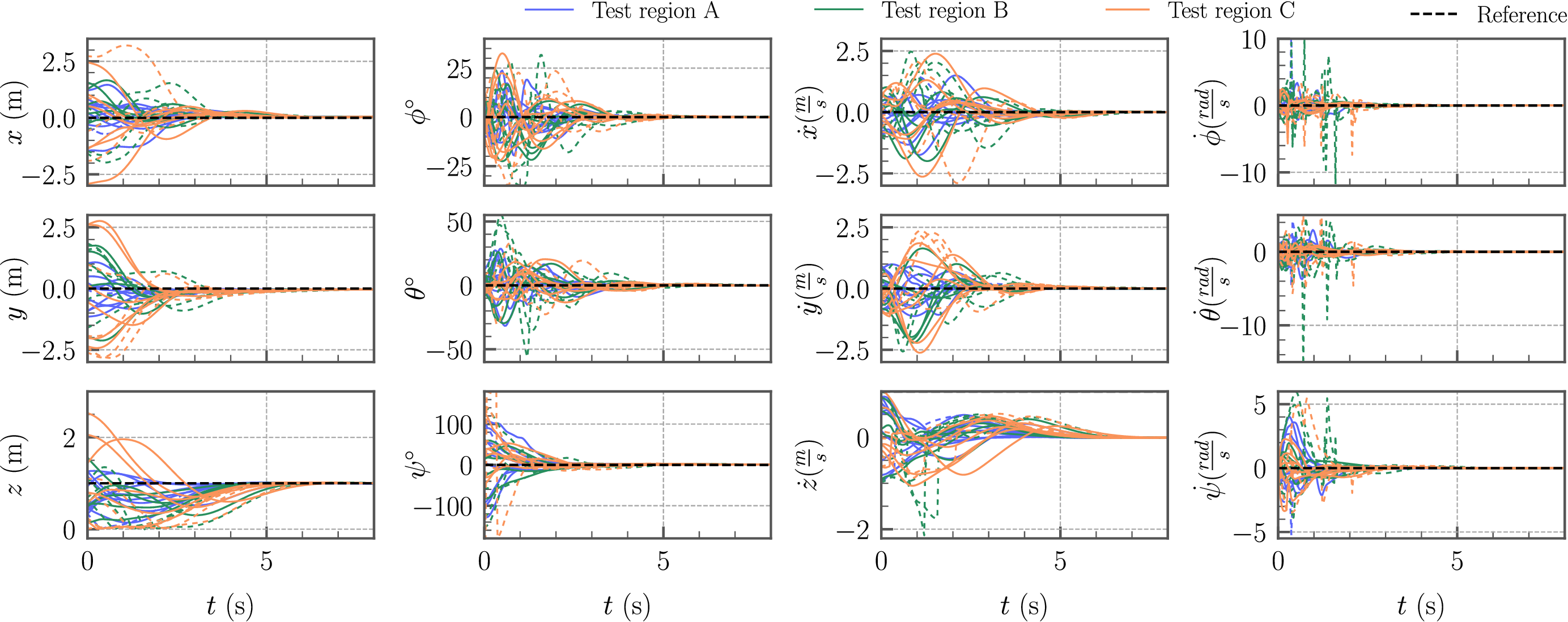}
    \caption{Performance of the CL-trained policy in achieving robust stabilization from 30 randomized initial conditions. Trajectories with initialization in the three regions A, B and C are shown in different colors: inner cylinder (Region A, blue), annular region (Region B, green), and outer cylinder (Region C, orange). Dashed traces indicate large transients, where the drone reaches close to the ground (seen from the evolution of the $z$ state), to overcome large initial velocities. The black dashed line represents the reference target values. The curriculum-trained policy consistently drives the system to the target position, from a diverse set of initial states.}
    \label{fig:robust_stabilization_variables}
\end{figure*}
%
%The plots show the evolution of the position (left column), orientation (left middle column), linear velocities (right middle column), and angular velocities (right column) of the Quadrotor.
\begin{figure}[h!]
    \centering
    \includegraphics[width=\linewidth]{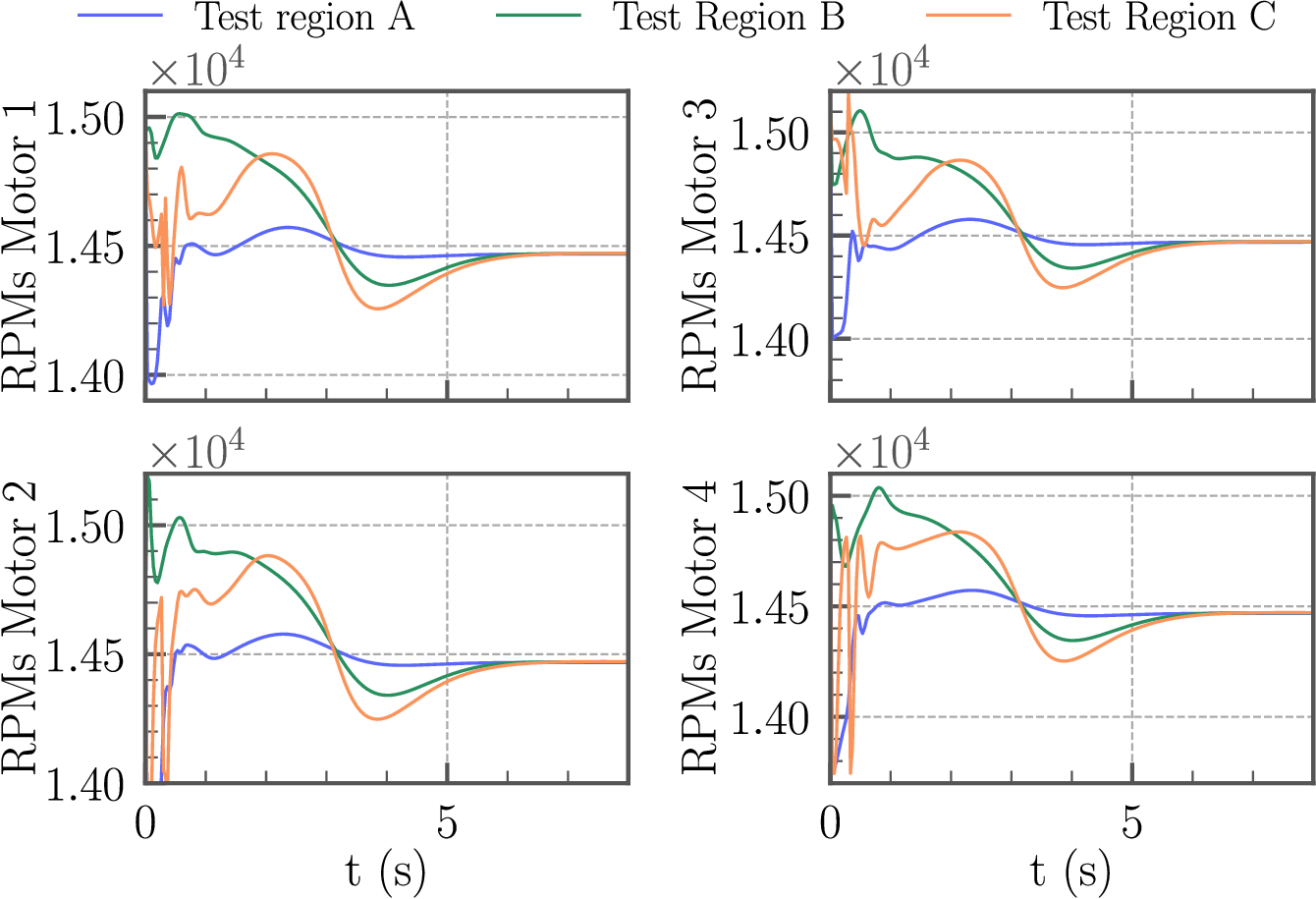}
    \caption{RPMs commanded by the curriculum-trained policy for the four motors of the Crazyflie Quadrotor, when the Quadrotor is initialized in the regions A, B and C.}
    \label{fig:rpms}
\end{figure}

\noindent \textbf{i) Robust stabilization:} \label{sec:robust_stabilization}
The performance of the CL-trained policy was evaluated by randomly setting the initial states of the drone and testing its ability to achieve the target configuration  $(x,y,z,\phi,\theta,\psi)=(0,0,1,0,0,0)$ and $(\dot{x},\dot{y},\dot{z},\dot{\phi},\dot{\theta},\dot{\psi)}=$$(0,0,0,0,0,0)$. 
We conducted 30 trials, each lasting 8 seconds, starting from randomized initial positions, orientations, linear velocities, and angular velocities. The 30 initial conditions were split among three regions shown in Figure \ref{fig:robust_stabilization}: Test region A, which is a cylinder with radius 1.5 m and height 1.5 m; Test region B, which is an annular region with inner radius of 1.5 m, outer radius of 2 m, and height of 2 m; Test region C, which is a cylinder with a radius of 2 m and height of 2 m. For all the trials, the initial roll ($\phi$) and pitch ($\theta$) angles were randomized in the range $[-15^\circ, 15^\circ]$ and the yaw ($\psi$) angle in the range of $[-180^\circ, 180^\circ)$, the linear velocities were initialized in the range of $[-1, 1] \frac{m}{s}$ and the angular velocities in the range of $[-1, 1] \frac{rad}{s}$. In all scenarios, as can be seen in Figure \ref{fig:robust_stabilization}, the drone successfully reached the target position, demonstrating the capability of the curriculum-trained policy in robustly stabilizing the Quadrotor at the target position. 

In Figure \ref{fig:robust_stabilization_variables}, we plot the evolution of the Quadrotor's position, orientation, linear velocity, and angular velocity from all 30 initial conditions. In most cases, the policy successfully negates the effect of non-zero initial linear and angular velocities and drives the Quadrotor to the target in less than five seconds, with smooth and safe transients. However, in five tests where the initial velocity was excessively high or the quadrotor was oriented away from the target, the drone exhibited complex and large transients, bringing the Quadrotor close to the ground, before driving it to the target configuration. In all 30 tests, the curriculum-trained policy consistently guided the drone to the target position with high accuracy, thus achieving robust stabilization of the Quadrotor from a diverse set of initial conditions. Figure \ref{fig:rpms} presents the motor RPMs generated by the curriculum-trained policy for three representative initial conditions from regions A, B, and C. In all three cases, the RPMs of all motors reach the steady-state hover RPMs.
%
%
% \noindent \textbf{Disturbance rejection:} \label{sec:disturbance_rejection}
% During a 40-second test run, the drone was subjected to external disturbances, to assess its ability to recover and maintain stability at the target position. Figure \ref{fig:disturbance_rejection} demonstrates that the policy successfully recovers from disturbances and returns to the desired position and attitude within approximately three seconds of the disturbances being introduced. The disturbances were applied manually by pushing the drone from its stable position at the target location, using the feature provided by Gym-PyBullet-Drones training and simulation framework.
%

\noindent \textbf{ii) Inspection Scenario:} A test scenario inspired by aerial inspection missions was designed to assess the robustness of the CL-trained policy in achieving a sequence of inspection view poses (target positions with target yaw angles). In this scenario, the drone is required to reach three columns placed at different locations and inspect points at different altitudes, and then move on to inspect a circular wall while maintaining its orientation aligned towards the wall by adjusting its yaw orientation. Although the CL-trained policy achieved zero yaw-orientation in all robust stabilization tests, it was observed that it robustly tracks non-zero yaw-reference angles only from the set $[-25^\circ, 25^\circ]$.  
In this scenario, using the CL-policy, the drone achieved each inspection target point with reference yaw orientation chosen from the set $[-25^\circ, 25^\circ]$, exhibiting performance that aligns with the predefined performance specifications set in the problem definition (Section \ref{sec:problem_statement}) of less than 5 seconds of settling time, less than 2.5 cm in positional steady state error, and less than $2^\circ$ error in yaw-tracking, while safely maintaining the roll and pitch angles between the safety boundaries of $15^\circ$ (set in training to enable smooth movements).
\begin{figure*}[h!]
    \centering
    \includegraphics[width=\linewidth]{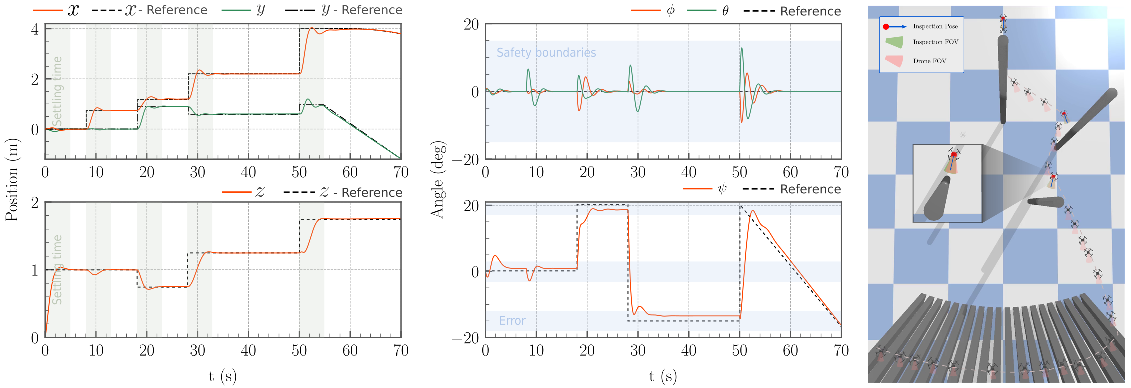}
    \caption{Over 70 s, the drone inspects three columns at different altitudes and yaw angles, then follows a trajectory parallel to a circular wall, inspecting it while continuously varying its yaw.}
    % \caption{\blue{During 70 seconds, the drone inspects three columns at different altitudes and yaw orientations and then follows a trajectory parallel to a circular wall, inspecting the wall, which requires varying its yaw orientation constantly.}}
    \label{fig:inspection}
\end{figure*}
%
%\section{Discussion}
%\noindent \textbf{Training process}. The implementation of the PPO algorithm followed the standard implementation provided by the Stable-Baselines3 library. 

%\noindent \textbf{Results}. 
The results presented so far demonstrate that the proposed curriculum learning approach, combined with the compounded reward function, allows the RL agent to achieve a high degree of stability, accuracy and robustness under complex initial conditions, including random linear and angular velocities. This suggests that the CL-trained agent has the potential to function as a reliable, general-purpose low-level controller for Quadrotors, subject to bridging the gap between simulation and real-world (sim2real) deployment.

\section{Conclusion and Future Directions}
This work proposed a sample-efficient three-stage curriculum learning methodology for efficiently and effectively training a reinforcement learning agent to accomplish robust stabilization for Quadrotors. An additive reward function was proposed to incorporate transient performance and steady-state accuracy requirements. From the results, it was concluded that curriculum learning was significantly more sample efficient compared to one-stage training (for the considered task and reward structure), which failed to achieve the learning objective even after 100 million time steps, with the same reward and PPO hyperparameters. The performance of the curriculum-trained RL policy was thoroughly tested in a physics-based simulator, from different initial conditions and in an inspection scenario requiring the policy to track a sequence of inspection view poses. 
%The results revealed that curriculum training is more sample-efficient than one-stage training, while yielding superior performance. 
%
The future work involves automating the curriculum setup and transferring the policy to real drones, thereby closing the simulation-to-reality gap. 
\bibliographystyle{IEEEtran}
\bibliography{references, referencesacc}
\end{document}